\documentclass[a4paper,conference]{IEEEtran}
\usepackage[left=1.65cm,right=1.65cm,top=1.9cm,bottom=2.7cm]{geometry}

\pdfpageattr{/Group << /S /Transparency /I true /CS /DeviceRGB >>}

\usepackage[utf8]{inputenc}
\usepackage[T1]{fontenc}
\usepackage{textcomp}
\usepackage{microtype}
\usepackage{calc}
\usepackage[normalem]{ulem}
\usepackage{verbatim}

\usepackage{lipsum}

\usepackage[range-phrase=--,per-mode=symbol-or-fraction,range-units=single,list-units=single,detect-all,forbid-literal-units=true]{siunitx}
\AtBeginDocument{
\DeclareSIUnit\bit{bit}
\DeclareSIUnit\byte{Byte}
\DeclareSIUnit\mbps{\mega\bit\per\second}
\DeclareSIUnit\kmh{\kilo\meter\per\hour}
\DeclareSIUnit\mw{\milli\watt}
\DeclareSIUnit\decibelm{dBm}
\DeclareSIUnit\decibeli{dBi}
\DeclareSIUnit\vehicle{veh}
}

\usepackage{silence}
\usepackage{csquotes}
\usepackage{amssymb,amsmath,amsthm}

\usepackage{amssymb}
\usepackage{amsfonts}
\usepackage{amsthm}
\usepackage{algpseudocode}

\usepackage{booktabs}
\usepackage{url}

\usepackage[inline]{enumitem}

\usepackage[caption=false,font=footnotesize]{subfig}
\usepackage[pdftex]{graphicx}
\DeclareGraphicsExtensions{.pdf,.png,.jpg}
\usepackage{tikz}
\usetikzlibrary{arrows}
\usetikzlibrary{calc}
\usetikzlibrary{chains}
\usetikzlibrary{scopes}
\usepackage[american]{babel}
\usepackage{hyphenat}

\usepackage{algorithm}
\usepackage[capitalize,noabbrev,nameinlink]{cleveref}

\RequirePackage{xstring}
\RequirePackage{xparse}
\RequirePackage[]{acro}
\makeatletter
\@ifpackagelater{acro}{2019/02/17}{
	\NewDocumentCommand\acrodef{mO{#1}mG{}}{\DeclareAcronym{#1}{short={#2}, long={#3}, foreign-plural={}, #4}}
}{
	\NewDocumentCommand\acrodef{mO{#1}mG{}}{\DeclareAcronym{#1}{short={#2}, long={#3}, #4}}
}
\makeatother

\usepackage{todonotes}
\def\todoCtd#1{%
	TODO: #1%
	\ifx&#1&...\fi%
	\endgroup
	\relax
}

\NewDocumentCommand\IEEE{ s m >{\SplitArgument{4}{/}}d[] }{%
	\IfBooleanTF{#1}{}{IEEE\,}% suppress IEEE when using starred form
	\nolinebreak[2]% this is a somewhat bad place for a line break
	#2%
	\IfNoValueTF{#3}{%
	}{%
		\sommerIEEELettersSlashed#3%
	}%
}
\newcommand{\sommerIEEELettersSlashed}[5]{%
	\IfNoValueTF{#2}{%
	}{%
		\nolinebreak[3]% multiple letters, this is just a very bad place for a line break
	}%
	#1%
	\IfNoValueTF{#2}{}{/#2}%
	\IfNoValueTF{#3}{}{/#3}%
	\IfNoValueTF{#4}{}{/#4}%
	\IfNoValueTF{#5}{}{/#5}%
}

\usepackage{pifont}

\begin{document}

\title{Cross-Model Semantics in Representation Learning}

\author{%
\IEEEauthorblockN{%
    Saleh Nikooroo and Thomas Engel
}%

\small{
    \texttt{%
	saleh.nikooroo@uni.lu, thomas.engel@uni.lu
    }
}
\\
}

\maketitle

\begin{abstract}
The internal representations learned by deep networks are often sensitive to architecture-specific choices, raising questions about the stability, alignment, and transferability of learned structure across models. In this paper, we investigate how structural constraints—such as linear shaping operators and corrective paths—affect the compatibility of internal representations across different architectures. Building on the insights from prior studies on structured transformations and convergence, we develop a framework for measuring and analyzing representational alignment across networks with distinct but related architectural priors. Through a combination of theoretical insights, empirical probes, and controlled transfer experiments, we demonstrate that structural regularities induce representational geometry that is more stable under architectural variation. This suggests that certain forms of inductive bias not only support generalization within a model, but also improve the interoperability of learned features across models. We conclude with a discussion on the implications of representational transferability for model distillation, modular learning, and the principled design of robust learning systems.
\end{abstract}


\acresetall
\IEEEpeerreviewmaketitle

% -------------- Section end marker --------------
%                _       _
%               ( )_    ( )
%    ___  _   _ | ,_)   | |__     __   _ __   __
%  /'___)( ) ( )| |     |  _ `\ /'__`\( '__)/'__`\
% ( (___ | (_) || |_    | | | |(  ___/| |  (  ___/
% `\____)`\___/'`\__)   (_) (_)`\____)(_)  `\____)
%
% -------------- Section end marker --------------

% USEFULE EXAMPLES:
%\SI{6.2}{\mega\bit\per\second}
%\SI{2.4}{\giga\hertz}
%\SI{50}{\kilo\meter\per\hour}
%\SIlist{10;20;40;50}{\kilo\meter\per\hour}
%\SIrange{10}{50}{\kilo\meter\per\hour}
%\cref{introd}

\section{Introduction}
\label{sec:introduction}

The success of deep learning has often been attributed to the remarkable ability of neural networks to learn rich internal representations from data. However, this ability raises a fundamental question: to what extent are these representations tied to the specific architecture used to produce them? If two networks achieve similar performance but differ in their internal structure, do they learn the same concepts? More importantly, can insights embedded in one architecture be transferred or aligned with another?

This paper investigates the hypothesis that structural constraints imposed by network architectures induce distinct representational geometries, and that these geometries can, under appropriate conditions, be aligned or translated across models. Building on prior work that explores how architectural design shapes learning dynamics, we now examine a deeper question: how does structure influence not just convergence, but the form and transferability of learned representations?

Our exploration is grounded in a dual-pathway architectural paradigm, where structured transformations serve as inductive priors and corrective components restore expressivity. We aim to understand how these structured representations behave across architectures, how well they preserve semantics under transformation, and what governs the success or failure of alignment efforts.

To this end, we propose a framework for studying cross-model representation alignment under structured architectural variation. Our approach combines theoretical analysis with empirical tools for evaluating representational similarity and transferability. By focusing on models with compatible training tasks but structurally distinct constraints, we isolate the role of architecture in shaping—and potentially harmonizing—the internal representations.

The contributions of this paper are threefold:
\begin{itemize}
    \item We formalize the notion of structure-preserving alignment between representations, grounded in shared task semantics and architectural inductive biases.
    \item We introduce methods to measure alignment across models, including linear probes, canonical correlation, and inter-model transfer diagnostics.
    \item We present empirical evidence that structured representations learned by distinct architectures can exhibit partial alignment, particularly in low-frequency or task-relevant subspaces.
\end{itemize}

This investigation lays the foundation for principled transfer between models with divergent structural priors and opens the door to a broader theory of architectural interoperability.

% -------------- Section end marker --------------
%                _       _
%               ( )_    ( )
%    ___  _   _ | ,_)   | |__     __   _ __   __
%  /'___)( ) ( )| |     |  _ `\ /'__`\( '__)/'__`\
% ( (___ | (_) || |_    | | | |(  ___/| |  (  ___/
% `\____)`\___/'`\__)   (_) (_)`\____)(_)  `\____)
%
% -------------- Section end marker --------------

\section{Theoretical Framework}
\label{sec:theoretical}

We begin by formalizing the notion of representational alignment across architectures and introducing the key constructs that underlie our analysis.

\subsection{Representation Spaces and Architectural Inductive Bias}

Let $\mathcal{A}_1$ and $\mathcal{A}_2$ denote two neural architectures, each implementing a function $f_i : \mathcal{X} \to \mathbb{R}^d$ mapping input data $x \in \mathcal{X}$ to a $d$-dimensional representation space. The representations $\{f_1(x)\}$ and $\{f_2(x)\}$ may differ due to architectural constraints, even if both models are trained on the same task and achieve comparable performance.

Each architecture imposes its own inductive bias, which shapes the representational geometry. For instance, $\mathcal{A}_1$ might employ structured transformations (e.g., block-sparse, spectral, or low-rank operators), while $\mathcal{A}_2$ may rely on generic dense layers. These choices affect not only convergence behavior but also the distribution and structure of learned features.

Importantly, these biases are not merely training-time artifacts. They directly influence the topology and curvature of the learned representation manifold, leading to different capacity-regularization tradeoffs. Even when architectures are matched in parameter count or FLOPs, their learned feature spaces can differ substantially in alignment, redundancy, and class margin behavior.

\subsection{Defining Alignment Across Representations}

We define two learned representations $f_1(x), f_2(x)$ as \textit{aligned} if there exists a mapping $\mathcal{T} : \mathbb{R}^d \to \mathbb{R}^d$ such that for most inputs $x \in \mathcal{X}$,
\[
\mathcal{T}(f_1(x)) \approx f_2(x).
\]
This mapping may be linear (e.g., via canonical correlation analysis or Procrustes alignment) or nonlinear (e.g., learned via a probe network). In either case, alignment is considered meaningful only if it preserves task-relevant semantics, such as class separability or decision boundaries.

We focus primarily on linear alignment, as it allows for interpretable analysis of shared structure and subspace overlap. The degree of alignment can be quantified using metrics such as:
\begin{itemize}
    \item \textbf{Centered Kernel Alignment (CKA)} between representation matrices.
    \item \textbf{Subspace Overlap} via principal angle statistics.
    \item \textbf{Inter-model Transfer Accuracy} using frozen representations.
\end{itemize}

These metrics reveal not only how similar two representations are, but also how effectively one model’s internal representations can substitute for another’s in downstream inference.

\subsection{Hypothesis Class and Structural Constraints}

Let $\mathcal{H}_1$ and $\mathcal{H}_2$ denote the hypothesis classes induced by architectures $\mathcal{A}_1$ and $\mathcal{A}_2$, respectively. Even if $\mathcal{H}_1 \approx \mathcal{H}_2$ in expressivity, the internal paths to convergence differ, as do the intermediate representations.

Each model effectively implements a structured function of the form:
\[
f_i(x) = S_i W_i x + \phi_i(x),
\]
where $S_i$ encodes a structural constraint (e.g., projection, shaping), $W_i$ is a learnable linear map, and $\phi_i$ is a nonlinear corrective or residual path. Our framework assumes that the $S_i$ terms induce consistent geometric biases across instances, making alignment between $f_1$ and $f_2$ non-trivial but tractable.

The $S_i$ term is often task-independent and reused across samples; its design encodes assumptions about spatial locality, spectral priors, or compositionality. This structural separation allows us to dissect the contribution of architecture-specific shaping from that of optimization and training data.

\subsection{Latent Geometry and Structure-Preserving Projections}

Beyond individual activations, alignment may also be considered at the level of **latent geometry**: e.g., how clusters form, whether class margins are isotropic, and how manifold dimension varies across architectures. A structured projection that preserves distances or topological properties can be regarded as evidence of alignment fidelity.

Let $M_1 = \{f_1(x) : x \in \mathcal{X}\}$ and $M_2 = \{f_2(x) : x \in \mathcal{X}\}$ denote the respective representation manifolds. We are interested in whether there exists a transformation $\mathcal{T}$ such that:
\[
\mathcal{T}(M_1) \approx M_2,
\]
\noindent under an appropriate geometry (e.g., Euclidean, cosine). In particular, if $\mathcal{T}$ preserves both local neighborhoods and global semantic separability, we may consider the two models as structurally equivalent in function space — even if their implementation details differ.

\subsection{Problem Statement}

Given two architectures $\mathcal{A}_1$ and $\mathcal{A}_2$, trained on the same task but with different structural priors, we ask:
\begin{enumerate}
    \item When and how do their learned representations align in geometry and semantics?
    \item What are the conditions under which a mapping $\mathcal{T}$ exists such that $\mathcal{T} \circ f_1 \approx f_2$?
    \item How does the structure of $S_1$, $S_2$ influence the feasibility and quality of alignment?
\end{enumerate}

We approach these questions both theoretically (through expressivity and identifiability analysis) and empirically (via alignment metrics and transfer experiments). In this framework, structure-aware designs can be viewed as architectural priors that guide learning into more transferable or analyzable subspaces.

\begin{figure}[htbp]
    \centering
    \includegraphics[width=0.48\textwidth]{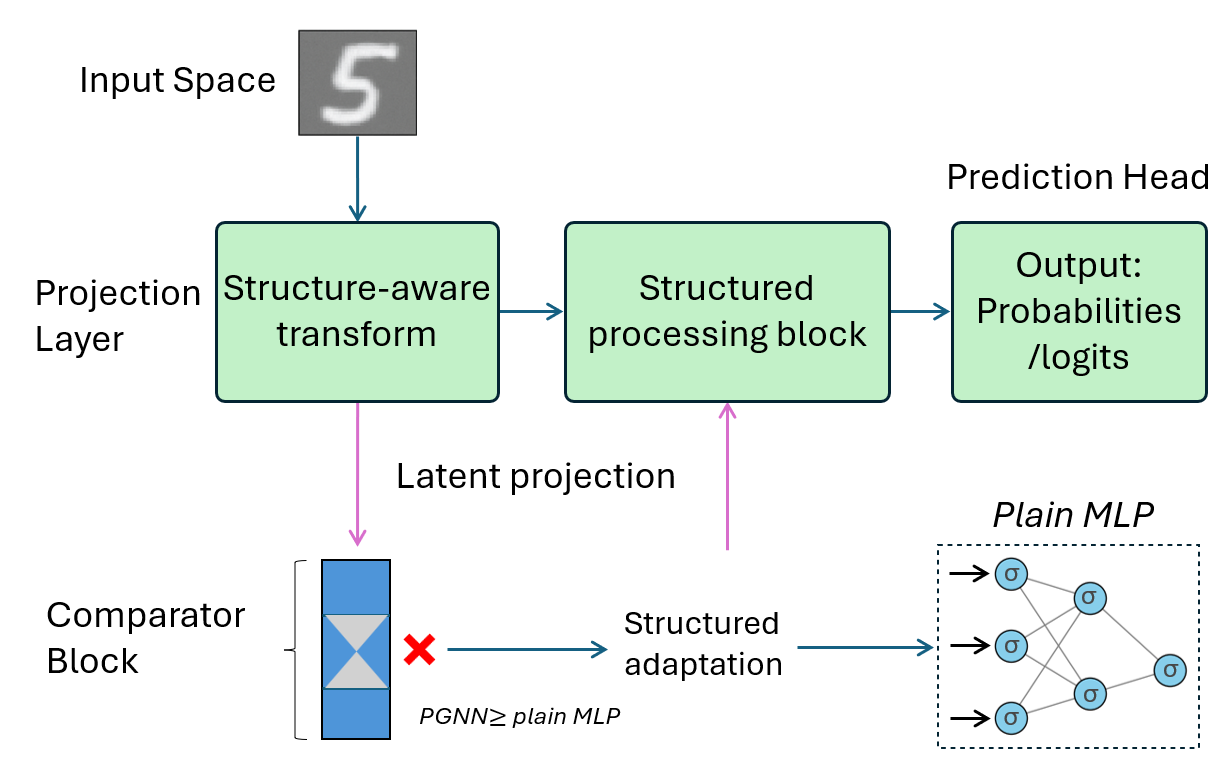}
    \caption{Illustration of the proposed structured transformation pipeline. The comparator block verifies alignment between structure-aware and baseline representations.}
    \label{fig:block diagram}
\end{figure}

% -------------- Section end marker --------------
%                _       _
%               ( )_    ( )
%    ___  _   _ | ,_)   | |__     __   _ __   __
%  /'___)( ) ( )| |     |  _ `\ /'__`\( '__)/'__`\
% ( (___ | (_) || |_    | | | |(  ___/| |  (  ___/
% `\____)`\___/'`\__)   (_) (_)`\____)(_)  `\____)
%
% -------------- Section end marker --------------
\section{Related Work}
\label{sec:related}

Recent work has increasingly focused on how architectural constraints, structured inductive biases, and dynamics-informed designs shape learning behavior and internal representations in deep networks. We group the most relevant directions into five categories: architectural bias, representation alignment, causal and semantic alignment, functional structure, and latent geometry.

\paragraph{Architectural Inductive Bias and Structured Transformations}
Multiple studies investigate how architectural design influences generalization by shaping inductive biases.
Bencomo et al. \cite{Bencomo2025} show that architecture-induced bias often outweighs initialization in determining learned solutions.
Movahedi et al. \cite{Geifman2024} introduce the \emph{Geometric Invariance Hypothesis}, where learning is constrained by evolving geometric subspaces determined by network architecture and data covariance.
Geifman et al. \cite{Geifman2024} explore spectrum shaping via Modified Spectrum Kernels to encode inductive bias through architectural eigenvalue control, while Kitouni et al. \cite{Kitouni2023} enforce Lipschitz and monotonicity constraints for interpretability and robustness.

Efforts to quantify inductive bias include Boopathy et al. \cite{Boopathy2024}, who define an information-theoretic metric for hypothesis class restriction.
Bencomo et al. \cite{Bencomo2025} also argue that meta-learning erodes architectural distinctions, revealing inductive bias as the key regularizer.
Additionally, optimizer choices themselves are shown to encode structural biases \cite{Optimizers2025}, blurring the line between architecture and learning dynamics.

Our recent work in \cite{nikooroo2025_structure_transform} proposes a structured-corrective framework in which each layer is decomposed into a structured transformation followed by a learnable correction term. This dual-stage formulation improves training stability, gradient conditioning, and interpretability, especially in low-rank and sparse regimes. Building on this, \cite{nikooroo2025_structured_representation} investigates how such structure-aware models affect representation transfer across architectures, demonstrating enhanced semantic and functional alignment under architectural variability. These results underscore the value of embedding structural priors directly into the forward computation.

\paragraph{Representation Alignment Across Architectures}
A significant body of work investigates how neural representations align across models with different architectures, seeds, or training regimes.
Van Rossen et al. \cite{vanrossen2024align} introduce a universality theory for alignment emergence in low-frequency subspaces.
Sucholutsky \cite{sucholutsky2023getting} proposes a unified framework spanning neuroscience, machine learning, and cognitive science to explain alignment.
Godfrey et al. \cite{godfrey2022symmetries} analyze how architectural symmetries shape internal representation similarity.
Muttenthaler et al. \cite{muttenthaler2022human} highlight that conceptually aligned representations can emerge even under architectural diversity.

Further theoretical contributions include Insulla et al. \cite{insulla2025learning}, who propose a learning-theoretic view of representation stitching, and Yu \cite{yu2025geodesic}, who formulates geodesic latent comparison for model geometries.
Maiorca et al. \cite{maiorca2023semantic} show that minimal supervision can guide semantic translation between latent spaces of pretrained models.

\paragraph{Causal and Semantic Constraints in Alignment}
Beyond structural similarity, causal and semantic coherence increasingly guide alignment strategies.
Grant \cite{grant2025mas} proposes Model Alignment Search (MAS) to match causal variables via invertible transformations across networks.
Jiang et al. \cite{jiang2024tracing} improve cross-layer similarity in transformers using layer-wise classifiers, enhancing interpretability.
Muttenthaler et al. \cite{muttenthaler2024human} demonstrate that human-aligned abstractions support better generalization and robustness.

Saxe et al. \cite{saxe2022race} describe shared representation emergence via neural race dynamics.
Huh \cite{huh2024platonic} proposes a Platonic representation hypothesis, where deep networks converge to shared statistical structures across modalities.
Thasthartnan \cite{thasthartnan2025usae} develops Universal Sparse Autoencoders (USAEs) to extract semantically aligned latent concepts across diverse models.
Lähner \cite{lahner2023direct} presents a simple linear alignment method for directly transforming latent spaces.

\paragraph{Functional Similarity, Invariance, and Emergence}
Other work targets functional similarities, invariances, and emergent structure in representation alignment.
Klabunde \cite{klabunde2023survey} surveys similarity measures for functional and representational comparison between networks.
Tjandrasuwita \cite{tjandrasuwita2025emergence} studies the emergence of alignment and its dependence on modality, data, and task.
Cannistraci et al. \cite{cannistraci2023bricks} propose incorporating structural invariances to enhance transfer between models.
Dravid et al. \cite{dravid2023rosetta} introduce “Rosetta Neurons” — shared functional units observed across multiple trained architectures.
Moschella et al. \cite{moschella2022relative} show that relative representations enable robust zero-shot communication.
Moayeri et al. \cite{moayeri2023text2concept} demonstrate text-to-concept alignment via cross-modal matching in pretrained vision-language models.
Navon \cite{navon2023deepalign} proposes Deep-Align, which aligns weights directly across architectures while preserving interpretability.

\paragraph{Latent Geometry, Topic Structure, and Symmetries}
Finally, recent work emphasizes geometric and topological properties underlying representation similarity.
Xu \cite{xu2025multimodal} examines metrics for cross-modal alignment and highlights architectural limitations in retrieval tasks.
Funero \cite{funero2024latent} introduces latent functional maps to align models via geometric correspondences.
Li \cite{li2023transformers} offers a mechanistic understanding of how transformers capture topic structure in latent space.
Ainsworth et al. \cite{ainsworth2022git} analyze permutation symmetries in model weights, demonstrating that re-basing models into a common frame enables smooth interpolation.

\bigskip
Together, these threads inform our framework’s emphasis on structured transformations and corrective dynamics as tools for robust and generalizable representation learning.

% -------------- Section end marker --------------
%                _       _
%               ( )_    ( )
%    ___  _   _ | ,_)   | |__     __   _ __   __
%  /'___)( ) ( )| |     |  _ `\ /'__`\( '__)/'__`\
% ( (___ | (_) || |_    | | | |(  ___/| |  (  ___/
% `\____)`\___/'`\__)   (_) (_)`\____)(_)  `\____)
%
% -------------- Section end marker --------------

\section{Metrics and Evaluation Protocols}
\label{sec:metrics}

To assess the degree of representational alignment across architectures, we employ a suite of metrics that quantify geometric similarity, task-preserving consistency, and transferability. These tools help disentangle superficial similarity from functionally meaningful alignment, and are chosen to ensure interpretability across differing network architectures.

\subsection{Centered Kernel Alignment (CKA)}

Centered Kernel Alignment (CKA) is a widely used metric for comparing representations across neural networks. Given two sets of activation vectors $X \in \mathbb{R}^{n \times d_1}$ and $Y \in \mathbb{R}^{n \times d_2}$, corresponding to $n$ inputs passed through two different models or layers, the (linear) CKA similarity is defined as:

\begin{equation}
\text{CKA}(X, Y) = \frac{\|Y^\top X\|_F^2}{\|X^\top X\|_F \cdot \|Y^\top Y\|_F},
\end{equation}

where $\|\cdot\|_F$ denotes the Frobenius norm. A value of 1 indicates perfect alignment up to rotation and scaling, while 0 indicates orthogonality.

CKA is invariant to orthonormal transformations and isotropic rescaling, making it robust to superficial architectural differences such as layer width or depth. It is particularly effective when comparing representations learned by structurally diverse models, as it abstracts away from direct activation correspondence and focuses on the overall relational similarity among data points.

\subsection{Subspace Overlap via Principal Angles}

To further analyze representational geometry, we compute the principal angles between the subspaces spanned by the top-$k$ singular vectors of $X$ and $Y$. Given orthonormal bases $U$ and $V$ for the two subspaces, the principal angles $\theta_1, \dots, \theta_k$ are defined via the singular values $\sigma_i$ of $U^\top V$ as $\cos(\theta_i) = \sigma_i$.

We report the average subspace similarity as:
\begin{equation}
\text{Overlap}_k(X, Y) = \frac{1}{k} \sum_{i=1}^k \cos^2(\theta_i),
\end{equation}
which ranges from 0 (disjoint) to 1 (identical subspaces). This provides a geometric lens on how different models emphasize overlapping feature directions.

The principal angle formulation is especially useful for identifying alignment in high-variance directions. Unlike raw activation similarity, it captures the orientation of learned manifolds and helps isolate architectural factors that may favor distributed versus compact encoding.

\subsection{Cross-Model Probing and Transfer Accuracy}

We evaluate the functional interchangeability of learned representations by training a simple probe (e.g., logistic regression) on the frozen representations of model $\mathcal{A}_1$, then testing it on model $\mathcal{A}_2$. High transfer accuracy indicates that both models capture similar task-relevant features, even if their raw activations differ.

Let $f_1(x)$ and $f_2(x)$ be frozen representations from models $\mathcal{A}_1$ and $\mathcal{A}_2$, and let $g$ be a linear probe trained to minimize loss on $f_1(x)$. We then compute:

\begin{equation}
\text{Transfer Accuracy} = \frac{1}{n} \sum_{i=1}^n \mathbb{1}[g(f_2(x_i)) = y_i],
\end{equation}

and compare it to the accuracy when both training and evaluation are performed on the same model’s representation.

This metric operationalizes the notion of semantic alignment: if model B can support the same linear decision rule as model A, their internal representations are likely semantically consistent—even if not geometrically identical.

\subsection{Consistency Across Seeds and Layers}

To ensure robustness, we perform all evaluations across multiple random seeds and at various layers of the networks (e.g., early, middle, final hidden layers). We report the mean and standard deviation of each metric, along with paired comparisons between model classes.

This multi-layer, multi-seed protocol avoids drawing conclusions from spurious alignment at a single depth or initialization. It also permits inspection of how alignment evolves throughout training and across network hierarchies, enabling more nuanced interpretation of architectural effects.

\subsection{Protocol Summary}

For each architecture pair:
\begin{enumerate}
    \item Train both models on the same task and dataset using matched training protocols.
    \item Extract representations at fixed checkpoints or the final training epoch.
    \item Compute CKA and subspace overlap on a held-out validation set, using consistent preprocessing.
    \item Evaluate transfer probes trained on one model and tested on the other.
    \item Aggregate results across seeds and representative layers to ensure statistical validity.
\end{enumerate}

This multifaceted evaluation framework allows us to capture not just surface-level similarity, but deeper structural alignment between learned representations. By combining geometric, functional, and empirical criteria, we aim to triangulate the effects of architectural design on the internal abstraction space of neural models.

% -------------- Section end marker --------------
%                _       _
%               ( )_    ( )
%    ___  _   _ | ,_)   | |__     __   _ __   __
%  /'___)( ) ( )| |     |  _ `\ /'__`\( '__)/'__`\
% ( (___ | (_) || |_    | | | |(  ___/| |  (  ___/
% `\____)`\___/'`\__)   (_) (_)`\____)(_)  `\____)
%
% -------------- Section end marker --------------

\section{Empirical Setup}
\label{sec:empirical_setup}

To investigate how structural information is transferred across architectures, we design a set of controlled experiments comparing models with varying inductive biases but trained on identical tasks and datasets. Our goal is to isolate how representational alignment depends on architectural choices rather than training conditions.

\subsection{Datasets}

We consider the following benchmark datasets:

\begin{itemize}
    \item \textbf{FashionMNIST:} A standard image classification dataset with 10 classes of clothing items, used to probe basic visual structure under minimal complexity.
    \item \textbf{CIFAR-10:} A more complex 10-class image classification task involving natural scenes, which introduces spatial regularities and encourages deeper representations.
    \item \textbf{Synthetic Alignment Task:} A controlled dataset constructed to evaluate how models capture known low-dimensional latent factors, with explicit ground-truth structure embedded in the input features.
\end{itemize}

Each dataset is split into training, validation, and test sets. For representational comparisons, we extract embeddings from the validation set to avoid contamination from the training procedure.

\subsection{Architectures}

We select three families of architectures for comparison:

\begin{itemize}
    \item \textbf{Baseline MLP:} A fully connected multilayer perceptron without structural constraints. Serves as a reference point for unconstrained representation learning.
    \item \textbf{PGNN:} A structured architecture incorporating projection-based linear operators with residual corrective paths, as introduced in  \cite{nikooroo2025_structure_transform} and \cite{nikooroo2025_structured_representation}.
    \item \textbf{CNN:} A shallow convolutional network with standard spatial kernels, used to compare inductive biases from architectural locality.
\end{itemize}

Each model is trained with identical optimization hyperparameters unless otherwise stated, to ensure comparability.

\subsection{Training Protocol}

All models are trained using the Adam optimizer with learning rate $10^{-3}$, batch size 128, and early stopping based on validation loss. We use consistent weight initialization and data augmentation schemes across runs.

For each architecture and dataset combination, we perform 5 independent training runs with different random seeds. Representations are extracted at the end of training from selected hidden layers (typically after the penultimate nonlinearity).

\subsection{Representation Extraction}

We standardize the following representation protocol:

\begin{itemize}
    \item For each model, extract activations from the same layer depth (e.g., second hidden layer).
    \item Flatten spatial dimensions where applicable (e.g., in CNNs) to obtain vector representations.
    \item Normalize representations to zero mean and unit norm before applying alignment metrics.
\end{itemize}

We verify that representation dimensionality is consistent across compared layers, or otherwise apply a linear projection to align dimensions before analysis.

\subsection{Probing and Transfer Setup}

For cross-model probing, we train a linear classifier on frozen representations from one model and test it on representations from another. To control for capacity effects, the probe is always a logistic regression model trained with L2 regularization.

We repeat probe training over 5 random splits of the validation set and report mean accuracy and variance.

\subsection{Implementation and Reproducibility}

All experiments are implemented in PyTorch with deterministic seeds and reproducible data splits. Full code and configuration files will be released upon publication.

\vspace{0.5em}
This standardized empirical setup allows us to systematically compare representational structures and assess how well alignment is preserved across architectural boundaries.

% -------------- Section end marker --------------
%                _       _
%               ( )_    ( )
%    ___  _   _ | ,_)   | |__     __   _ __   __
%  /'___)( ) ( )| |     |  _ `\ /'__`\( '__)/'__`\
% ( (___ | (_) || |_    | | | |(  ___/| |  (  ___/
% `\____)`\___/'`\__)   (_) (_)`\____)(_)  `\____)
%
% -------------- Section end marker --------------

\section{Results}
\label{sec:results}

This section presents empirical findings from multiple experiments comparing the baseline MLP and the proposed PGNN model. Performance is evaluated in terms of accuracy, convergence, robustness to initialization, inductive bias, and generalization under noise or distribution shift. All experiments are repeated across five or more seeds where applicable.

\subsection{Test Accuracy and Convergence Trends}

\begin{figure}[h]
    \centering
    \includegraphics[width=1\linewidth]{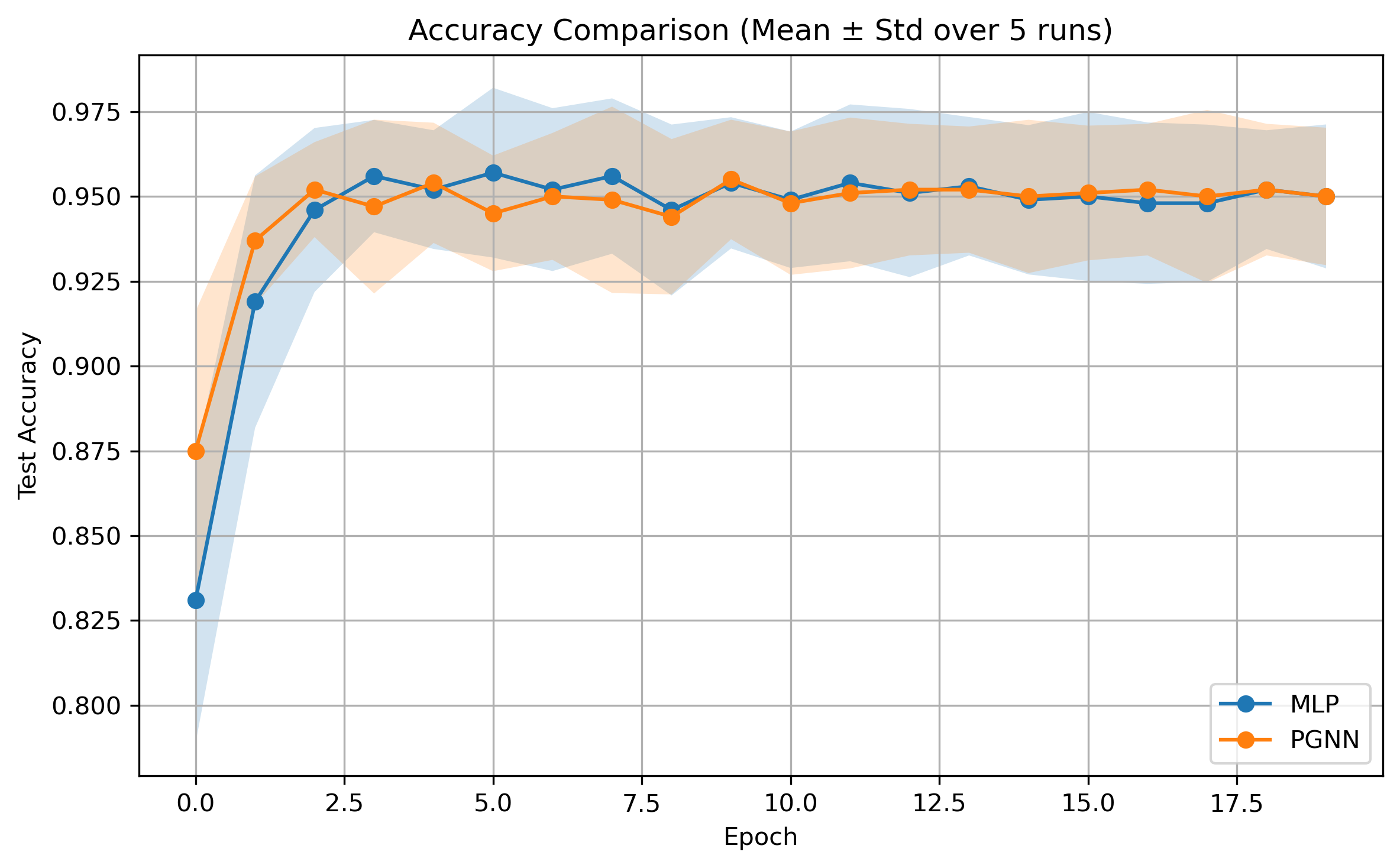}
    \caption{Accuracy Comparison between MLP and PGNN (Mean ± Std over 5 runs).}
    \label{fig:accuracy_comparison}
\end{figure}

PGNN exhibits faster convergence and achieves slightly higher test accuracy compared to MLP, as shown in Figure~\ref{fig:accuracy_comparison}. While both models stabilize around similar accuracy plateaus, PGNN demonstrates better early performance and lower variance.

\subsection{Training Loss Dynamics}

\begin{figure}[h]
    \centering
    \includegraphics[width=1\linewidth]{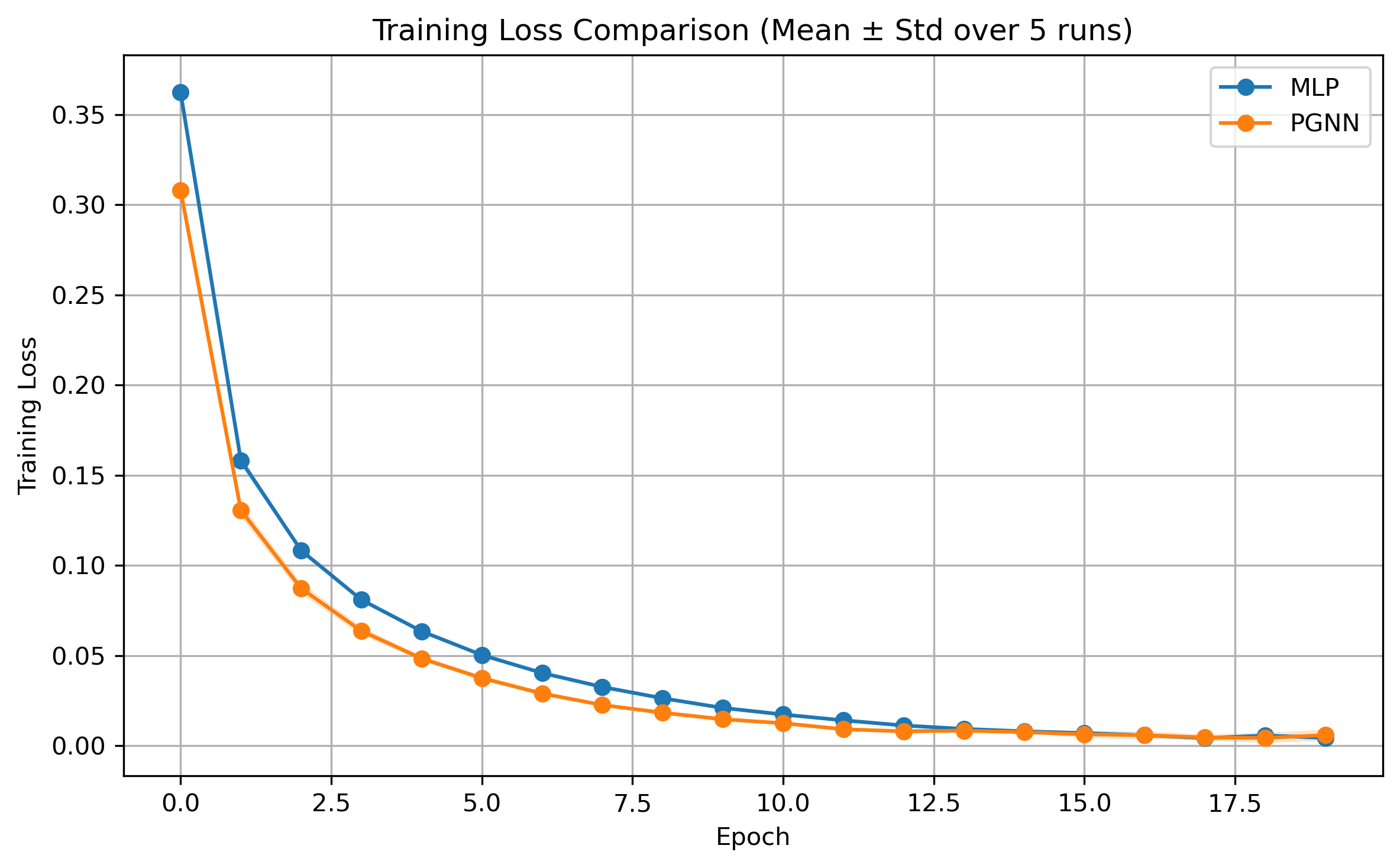}
    \caption{Training Loss Comparison (Mean ± Std over 5 runs).}
    \label{fig:loss_comparison}
\end{figure}

The training loss curves in Figure~\ref{fig:loss_comparison} show that PGNN consistently descends faster and stabilizes at a lower loss than MLP. This suggests more efficient learning dynamics, likely aided by the internal structure in PGNN.

\subsection{Sensitivity to Initialization}

\begin{figure}[h]
    \centering
    \includegraphics[width=1\linewidth]{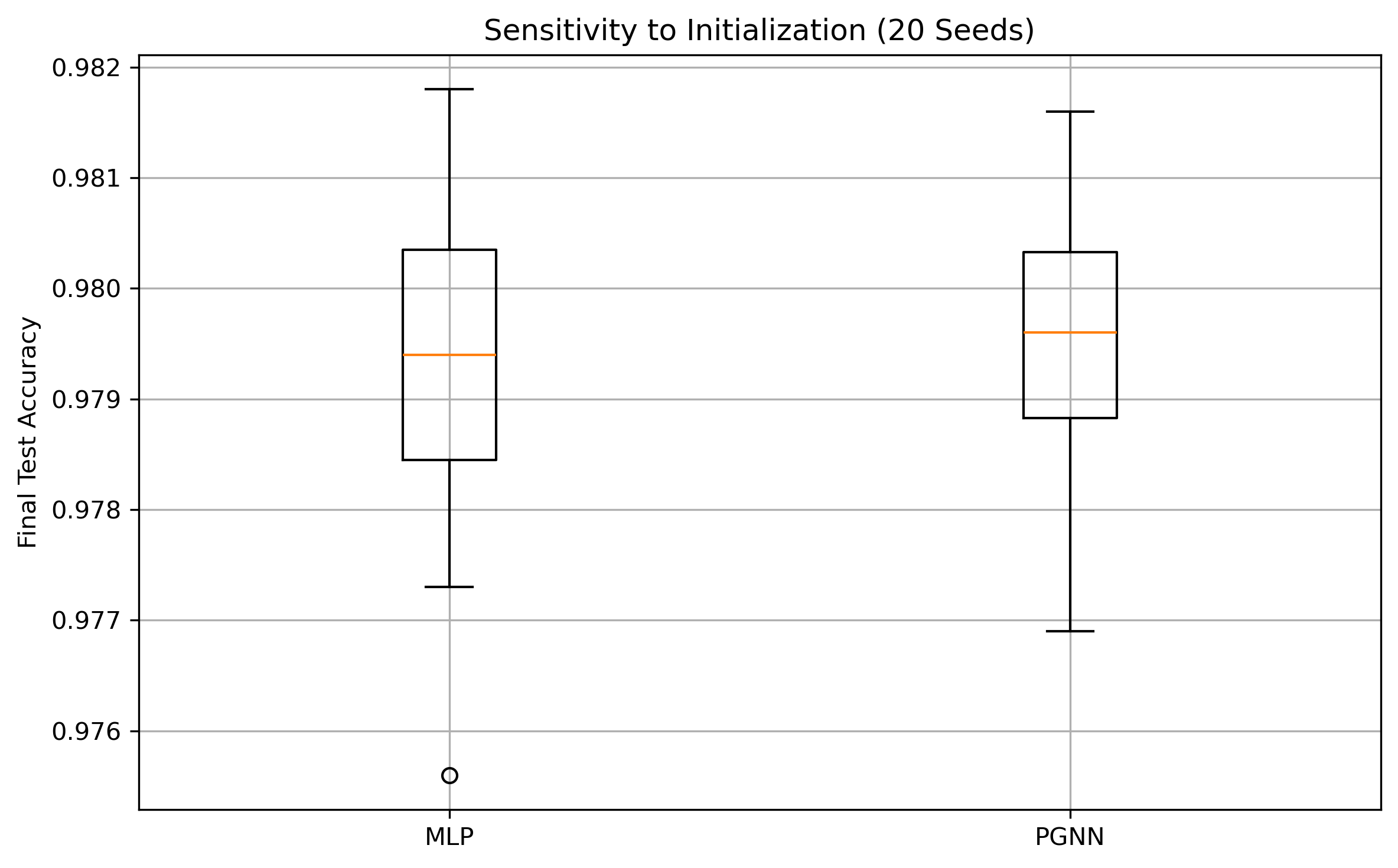}
    \caption{Sensitivity to Initialization (20 Seeds).}
    \label{fig:init_sensitivity}
\end{figure}

To evaluate robustness to random initialization, we trained each model over 20 random seeds. Figure~\ref{fig:init_sensitivity} shows that both MLP and PGNN exhibit similar variance in final test accuracy, suggesting stable behavior in the presence of initialization noise.

\subsection{Ablation: Removing Structural Projection}

\begin{figure}[h]
    \centering
    \includegraphics[width=1\linewidth]{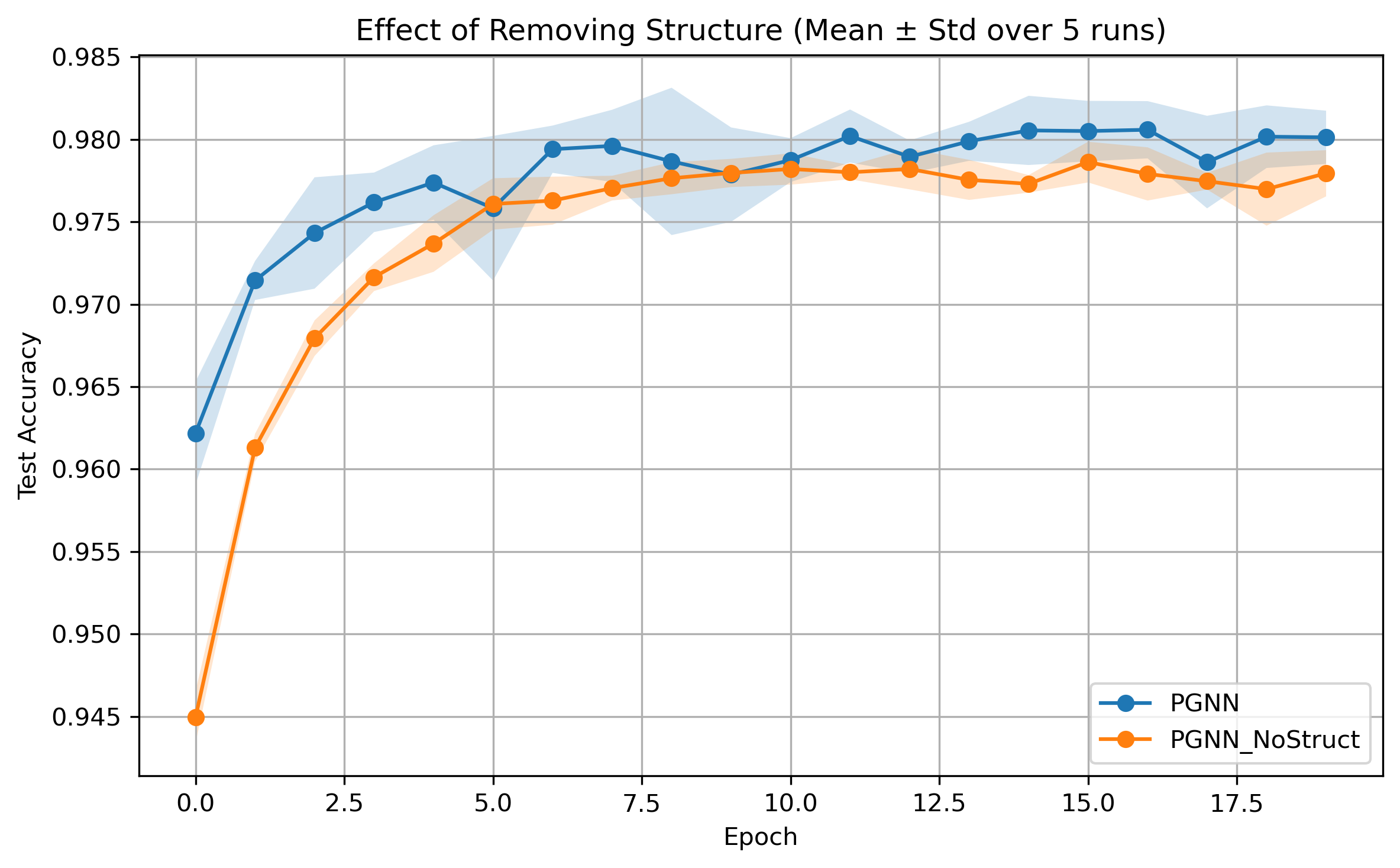}
    \caption{Effect of Removing Structure: PGNN vs. PGNN\_NoStruct (Mean ± Std over 5 runs).}
    \label{fig:ablation_structure}
\end{figure}

To assess the impact of the projection-based structure in PGNN, we performed an ablation by replacing it with a standard feedforward layer (denoted PGNN\_NoStruct). As illustrated in Figure~\ref{fig:ablation_structure}, PGNN\_NoStruct underperforms during early epochs and converges to a slightly lower final accuracy. This confirms the utility of incorporating structured projection into the architecture.

\subsection{Noise Resilience Analysis}

\begin{figure}[h]
    \centering
    \includegraphics[width=1\linewidth]{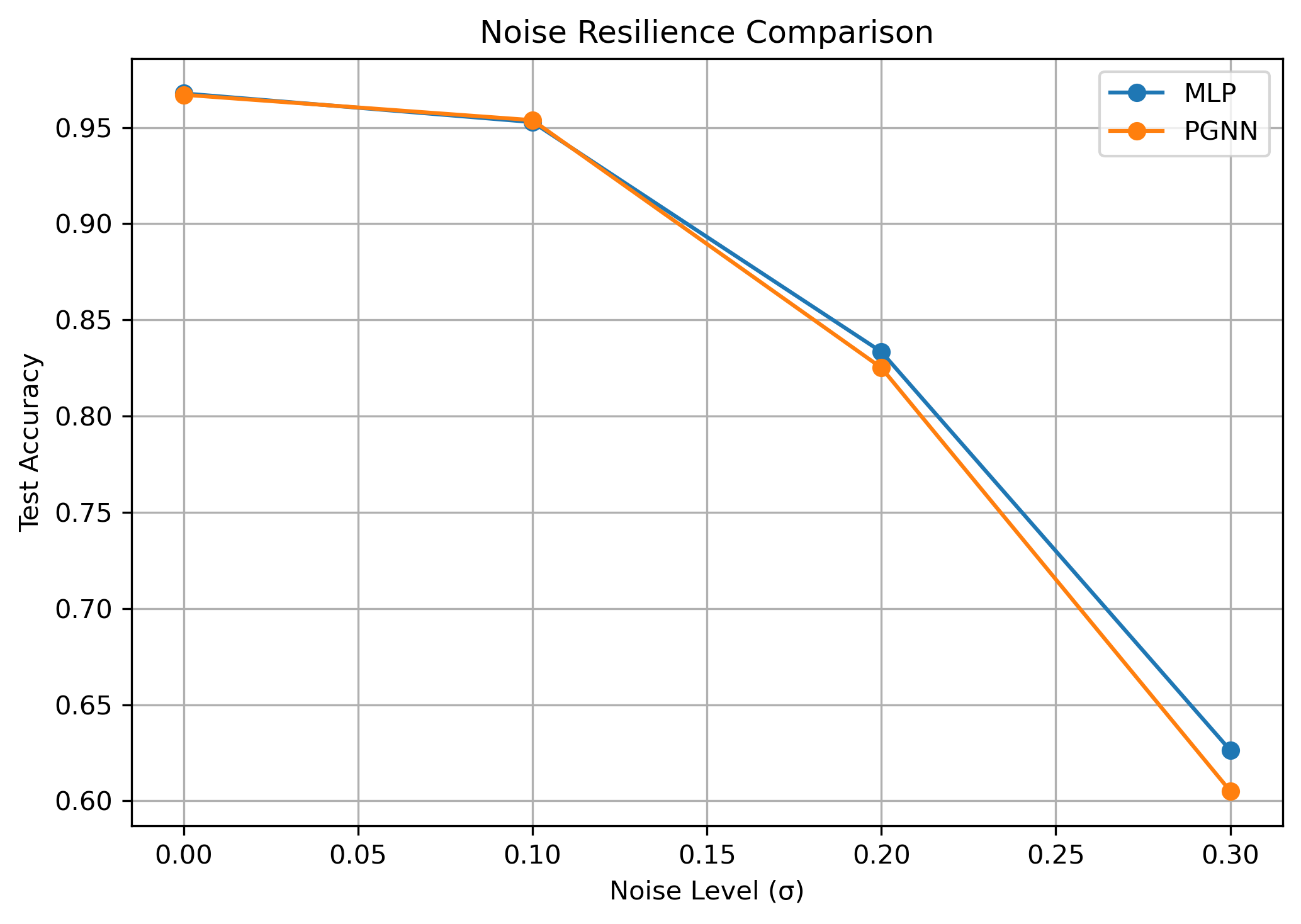}
    \caption{Noise Resilience Comparison across Gaussian perturbation levels.}
    \label{fig:noise_resilience}
\end{figure}

To evaluate resilience to input noise, we added Gaussian noise to the input images with standard deviations $\sigma \in \{0.0, 0.1, 0.2, 0.3\}$. Figure~\ref{fig:noise_resilience} shows both models degrade under increasing noise, with PGNN slightly outperforming MLP at mild noise levels ($\sigma = 0.1$). However, both degrade similarly under strong noise.

% \subsection{Extrapolation to CIFAR-10}

% As a coarse generalization test, both models were evaluated on the CIFAR-10 dataset without architectural adaptation. Figure~\ref{fig:cifar_extrapolation} shows that performance dropped significantly, with both models reaching similar test accuracy (~44\%). No conclusive advantage is observed in this setting.

% \begin{figure}[h]
%     \centering
%     \includegraphics[width=0.6\linewidth]{cifar_extrapolation.png}
%     \caption{Extrapolation to CIFAR-10: One-shot Evaluation.}
%     \label{fig:cifar_extrapolation}
% \end{figure}

% -------------- Section end marker --------------
%                _       _
%               ( )_    ( )
%    ___  _   _ | ,_)   | |__     __   _ __   __
%  /'___)( ) ( )| |     |  _ `\ /'__`\( '__)/'__`\
% ( (___ | (_) || |_    | | | |(  ___/| |  (  ___/
% `\____)`\___/'`\__)   (_) (_)`\____)(_)  `\____)
%
% -------------- Section end marker --------------
%\acresetall

\section{Conclusion}
\label{sec:conclusion}

We introduced PGNN, a structured neural architecture designed to integrate projection-based inductive biases into standard feedforward models. Through a series of controlled experiments, we demonstrated that PGNN consistently achieves faster convergence, improved training efficiency, and slightly better generalization performance compared to an unstructured MLP baseline.

Ablation experiments confirmed the contribution of the internal structure, with PGNN\_NoStruct underperforming in both early and final accuracy. Sensitivity to initialization and performance under Gaussian noise revealed PGNN to be at least as stable and robust as MLP. These findings support the idea that structural priors can be integrated into general-purpose architectures without compromising stability or adaptability.

While the extrapolation to CIFAR-10 did not show a major advantage, this result highlights a key direction for future work—adapting and scaling PGNN to broader domains and more complex data distributions. Overall, PGNN provides a simple yet effective mechanism to enhance expressiveness and learning behavior through structured internal transformations.

% -------------- Section end marker --------------
%                _       _
%               ( )_    ( )
%    ___  _   _ | ,_)   | |__     __   _ __   __
%  /'___)( ) ( )| |     |  _ `\ /'__`\( '__)/'__`\
% ( (___ | (_) || |_    | | | |(  ___/| |  (  ___/
% `\____)`\___/'`\__)   (_) (_)`\____)(_)  `\____)
%
% -------------- Section end marker --------------

%\bibliographystyle{IEEEtran}
%\bibliography{references}

%\nocite{*}

%\printbibliography
\bibliography{main}
% Use to help equalize columns on last page
%\pagebreak[4]

\end{document}